\documentclass{article}

\usepackage{arxiv}

\usepackage[utf8]{inputenc} 
\usepackage[T1]{fontenc}    
\usepackage{hyperref}       
\usepackage{url}            
\usepackage{booktabs}       
\usepackage{amsfonts}       
\usepackage{enumitem}
\usepackage{graphicx}
\usepackage{multicol}
\usepackage{float}
\usepackage[ruled,vlined]{algorithm2e}
\usepackage{indentfirst}
\usepackage{subcaption}
\usepackage{multirow}

\newcommand{\xsi}{x^{(i)}}
\newcommand{\ysi}{y^{(i)}}
\title{Motion-Based Handwriting Recognition }

\author{
  Junshen Kevin Chen \\
  Stanford University\\
  \texttt{jkc1@stanford.edu} \\
   \And
  Wanze Xie \\
  Stanford University\\
  \texttt{wanzexie@stanford.edu} \\
    \And
  Yutong (Kelly)  He \\
  Stanford University\\
  \texttt{kellyyhe@stanford.edu} \\
}

\begin{document}
\maketitle

\begin{multicols*}{2}

\section{Introduction}

It is prevalent in today’s world for people to write on a touch screen with a smart pen, as there is a strong need to digitize hand-written content, to make the review and indexing easier. However, despite the success of character recognition on digital devices \cite{bib1, bib2, bib3}, requiring a digitizer as the writing surface poses a possibly unnecessary restriction to overcome. In addition, in VR and AR applications, it is also hard to recognize the texts written with the motion sensors \cite{bib4, bib5}.

We define the problem for identifying handwriting texts given the motion data collected by a sensor as a \textbf{motion-based handwriting recognition problem}. Such problem is different from other approaches such as Optical Character Recognition (OCR)\cite{ocr} problem as it requires our model to not utilize any visual features of the written text. In this project, we focus on classifying individual characters with motion data. We propose a solution by using a pen equipped with motion sensor to predict the English letters written by the user. Instead of using image data or on-screen stroke data, we analyze the acceleration and gyroscopic data of the pen using machine learning techniques to classify the characters the user is writing.

To tackle this problem, we first collected our own dataset by building the hardware and inviting 20 different users to write lowercase English alphabet with the equipment. The input to our algorithm is the yaw, pitch and roll rotation values of the sensor collected when the user is writing with the pen. After preprocessing the data using feature interpolation, data augmentation, and denoising using an autoencoder, we use the processed data to experiment with 4 different classifers: KNN, SVM, CNN and RNN to predict the labels that indicate which letters the user was writing when the sensor data was collected.

Our contribution are as the following:
\begin{enumerate}[label=-]
    \item We define a new problem for recognizing handwriting text using only motion data.
    \item We built a hardware to collect a new dataset with 20 different subject writing lowercase English alphabet 20 times.
    \item We designed and implemented a pipeline that solves the defined problem with feature interpolation, data augmentation, denosing autoencoding and 4 different classifers.
    \item We conduct experiments for pipeline with two different settings and our best model achieves $86.6\%$ accuracy in the random split experiment and $53.6\%$ in the subject split experiment.
\end{enumerate}

All codes and dataset are available at our GitHub repository \textit{\href{https://github.com/RussellXie7/motion-based-handwriting-recognition}{https://github.com/RussellXie7/motion-based-hand writing-recognition}} and demonstrations are available at \textit{\href{https://www.youtube.com/watch?v=SGBSVo2U12s}{https://youtu.be/SGBSVo2U12s}}.

\section{Related Work}

\paragraph{Sensor-Based Gesture Recognition} Recently, there have been lots of researches for various ways of leveraging inertial motion unit (IMU) data to predict the gesture or the activity of users \cite{Kim_2019, ganguly2018kinect, 7418327, 6470686, 6208895}, but few studies make use of the IMU data to predict the handwriting letter due to the lack of relevant dataset. Oh et al. analyzed using inertial sensor based data to recognize handwritten arabic numbers handwritten in the 3D space \cite{1363896}. However, one major problem with the system described in the study is that it requires user to wave hand in the space to outline the trajectory of the intended number to write, which contradicts the writing habits of people in daily activity (\textit{e,g.}, write on the table, or write in space with pen-tip pointing down). Zhou et al. \cite{4601870} performed a similar study for sensor-based capital letter classification using a single layer unsupervised network. Our study shares the same spirit, but instead we have a more practical set-up of the handwriting sensing device and explore modern machine learning techniques to address both data pre-processing and classification tasks.

\paragraph{Vision-Based Handwriting Recognition} It has been very successful in using vision-based approaches to tackle the handwriting recognition problem, with OCR \cite{ocr} being one of the most prominent research field, which has been prosperous in applications like translating manuscript into digitized texts. Other vision based researches propose tracking the motion of human finger \cite{finger} or forearms \cite{940024} using RGB or depth camera. Nevertheless, without a digital screen, an OCR based approach often require an image of the handwritten manuscript, which is asynchronous between the writing and recognition event. Camera based approaches often require an extra layer of set-up in order to recognize and convert handwriting into digital texts in real time, and occlusions can often be a challenge as well \cite{Chen2008}. Our system aims to approach the problem with a simpler framework by only requiring a pen.

\section{Dataset and Features}

For this project, we built the hardware with a MPU9250 9-axis motion sensor, an Arduino Uno R3, and a 3D printed mount to connect stylus and the sensor. We also constructed a writing grid to help subjects write data in a consistent size.

\vspace{-4px}
\begin{figure}[H]
    \centering
    \includegraphics[scale=0.16]{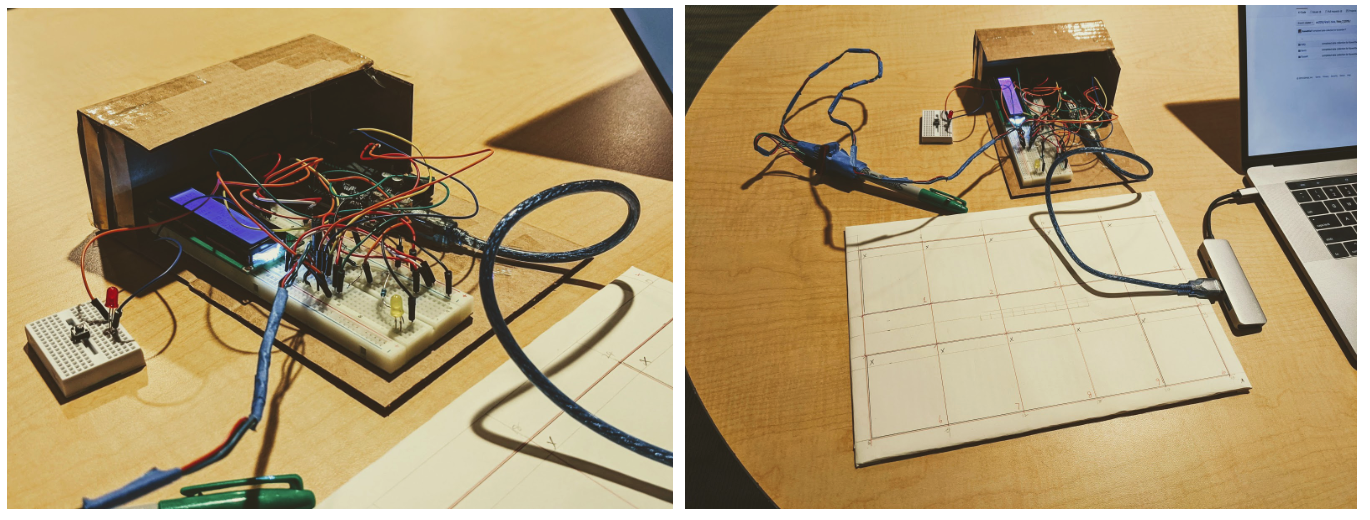}
    \caption{Hardware (left) and writing grid (right)}
    \label{fig:hardware}
\end{figure}
\vspace{-8px}

We formalize our data collection strategy by recording 20 writings of each lowercase letter from the alphabet from numerous subjects, and collected our own original data set. From the start of subject pressing down the record button, move the stylus to write one single letter, to the release of the button constitutes a \textbf{writing event}, producing data known as a \textbf{sequence}, consisting of as many rows (one row per frame) as the sequence takes. Then, some frames a sample of a writing sequence is in the following format:

\vspace{-5px}

\begin{table}[H]
\centering
\begin{tabular}{lllllll}
\hline
td & yaw    & pitch  & roll   & ax     & ay      & az      \\ \hline
7  & 90.10 & -10.34 & -20.02 & 206.9 & -374.1 & 1052.9 \\
25 & 90.27  & -9.86  & -20.29 & 193.0 & -401.7 & 1046.2 \\ \hline
\end{tabular}
\vspace{3pt}
\caption{Frames from a sample of writing sequence}
\label{tab:sample-sequence}
\end{table}

\vspace{-8px}

Where \textbf{td} is the time delta between the last frame and current frame sampled by the sensor in $ms$, \textbf{yaw, pitch, roll} are rotation values in $degrees$, and \textbf{ax, ay, az} are acceleration on each Cartesian axis in $mm/s^2$.

We have collected 20 sequences per letter from 20 subjects, resulting in a dataset of $10,400$ sequences.

\subsection{Visualization}

We build a visualizer to gain insight to the data before developing several techniques to process and use the data to train various models.

\begin{figure}[H]
    \centering
    \includegraphics[scale=0.25]{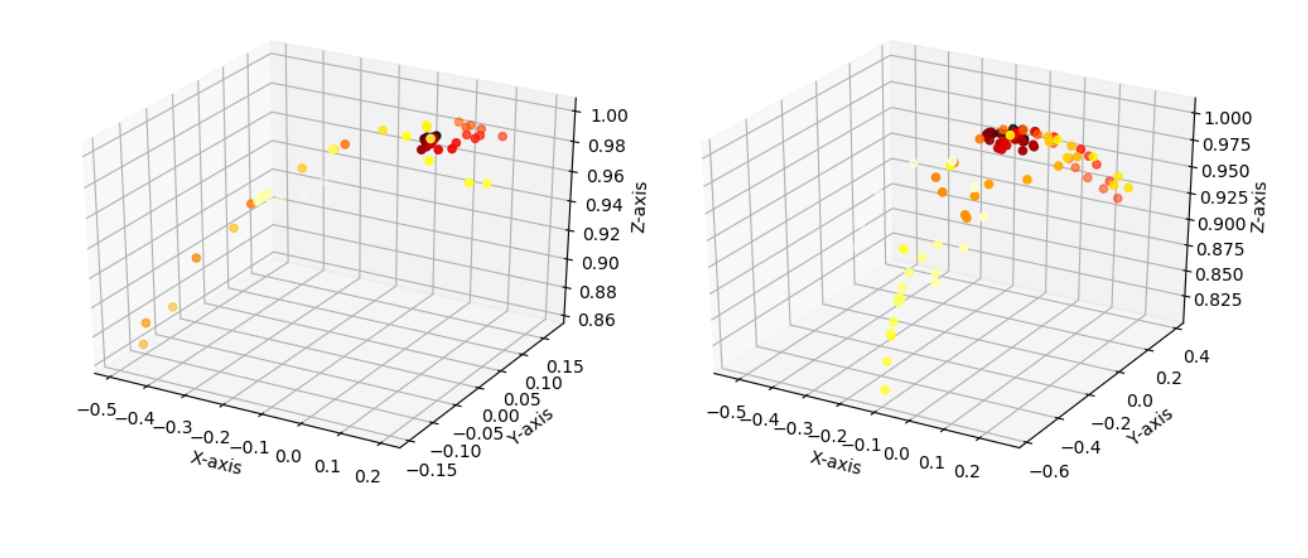}
    \caption{Letter 'a' written by two subjects}
    \label{fig:viz_a}
\end{figure}

\vspace{-10px}

The above diagram plots the same letter 'a' written by two different subjects, visualized by rotating a unit vector anchored to the origin, and plotting the tail of the vector. 



Observing the above, even though the traces of rotation values do not resemble the letter itself, mainly due to the sensor records the tail of the stylus, we notice that writing of the same letter share characteristics across subjects, and distinct letters have identifiable traces against each other.

\subsection{Data Augmentation}

Due to the difficulty to collect a large a mount of data within the time scope of this project, we implement a data augmentation pipeline to generate new samples by:
\begin{enumerate}
    \item Add a Gaussian noise centered at 0 and with customized small variance to each entry;
    \item Rotate the object by a small random quaternion vector;
    \item Stretch each sequence by three random scalar constants corresponding to yaw, pitch, roll.
\end{enumerate}

\subsection{Calibration and Normalization}

The sensor is not by default calibrated before each subject begin writing sequences. To solve this problem, we ask each subject to hold the stylus upright for 10 seconds, and use the mean of that sensor recording to subtract from all frames recorded from this subject. 

For rotation values, we want data to be invariant to different subjects' holding the stylus differently, and therefore we also subtract frame 0 of each sequence from all frames in this sequence, to get a delta rotation value.

\subsection{Data Interpolation and Re-sampling}

For some models that do not have the inherent structure for time series data, such as a CNN, we need to transform the input sequence to a flat vector before the network can be trained. There are two problems in merging and flattening all frames in each sequence directly:  (1) \textbf{subjects write in different speeds}, and therefore some writing of the same
letter may produce a longer sequence than others and hence produce vectors with different length; (2) \textbf{the sensor
samples in a variable rate}, and therefore the time delta between each timestamp varies.

To solve this problem, we design a procedure that normalizes sequences so that they have the same number of features. To obtain the desired fixed number of features $N$, where $N$ is ideally the average number of frames in each sequence, do:

\begin{enumerate}
    \item Extract the td, yaw, pitch, roll values from each data sequence and create a $3\times1$ array for them, exclude td;
    \item Map each yaw, pitch, roll to corresponding time stamps;
    \item Create a 1D linear interpolation model for each sequence;
    \item Generate linearly spaced time stamps based on the lower and upper bound of the original timestamps;
    \item Calculate the interpolated yaw, pitch, roll value for each timestamp generated in step 4;
    \item Pair up corresponding yaw, pitch, roll values and merge them into a single vector of length $3N$ if flatten, or of shape $N \times 3$ if not flatten;
\end{enumerate}

We formalize such feature extractor as: $f(x): \mathbf{R}^{M \times 4} \to \mathbf{R}^{3N} \cup \mathbf{R}^{N \times 3} $, $M$ is number of frames in each sequence that varies, and each frame contain $\{td,y,p,r\}$. Output is in either $\mathbf{R}^{3N}$ or $\mathbf{R}^{N \times 3}$ depending on if we choose to flatten the data.

\subsection{Denoising Autoencoder}

One of the limitations of our system is that it collects data at an unstable sample rate, and also because of the small vibration of the pen when people move their hands, the data we collected are noisy by nature. We explore a neural network based approach using autoencoder to deal with this problem.

An autoencoder is an architecture composed of both an encoder and a decoder and it is trained to minimize the reconstruction error between the encoded-then-decoded output and the input. Before training our classifiers, we can choose to apply autoencoder respectively to the raw yaw, pitch and roll data, and we analyze the results in Section 5.

The input and output of the autoencoder are both a vector in $\mathbf{R}^N$, where $N$ is the number of features. In all our experiments $N=100$. The hidden encoding and decoding layer both have 128 features and the encoded feature size is 64. All layers are fully connected, and we use mean-squared error defined as below to train the model.

\[
J(x, \hat{x}) = \frac{1}{N} \sum_{i=1}^N (\xsi - \hat{x}^{(i)})^2
\]
$\xsi \in \mathbf{R}$, which is the rotation value in degrees from one of the yaw, pitch and roll data and $\hat{x}^{(i)}$ is the reconstructed input.


As a result, with a encoded feature size smaller than the number of features of the input, the autoencoder can learn the most efficient way to represent the data and therefore produce a result with most noises excluded.

\section{Methods}

\subsection{K Nearest Neighbors (KNN)}

Intuitively, even though motion data does not construct a pattern similar to the letter itself, it should still be distinguishable, i.e. the nearest neighbors to an unseen ``a'' should be predominantly ``a''s.

Therefore we choose KNN \cite{knn} as a baseline classifier. For any unseen sample, choosing $K$ seen samples that are similar to it, then taking the most popular label to be the prediction. The algorithm is simple as there is no ``training'' involved.

\begin{algorithm}[H]
\SetAlgoLined
\caption{Predicting with K nearest neighbors}
\SetKwInOut{Input}{Input}\SetKwInOut{Output}{Output}
\Input{$\hat{x}$, an unseen sample; $K$, number of neighbors}
\Output{$\hat{y}$, the prediction}
\KwData{$\forall i, (\xsi, \ysi)$, seen samples and their labels}

$D^{(i)}\leftarrow  \|\xsi - \hat{x}\|$ for all $\xsi$ \\
$\hat{S} \leftarrow $  $K$ samples $(\xsi, \ysi)$ for the smallest value of $D^{(i)}$ \\
yield $\arg \max_{y} \sum_{(\xsi, \ysi) \in \hat{S}} \mathbf{1}[\ysi = y] $ \\
\end{algorithm}

\subsection{Support Vector Machine (SVM)}

We use a ``one-vs-all'' strategy to implement the multi-class SVM \cite{svm} since our data set is balanced (there are roughly equal number of samples for each label). For each of the 26 letters, train an SVM with polynomial kernel with positive samples are the ones of the corresponding letter, and negative samples are samples of all other letters.

\begin{algorithm}[H]
\SetAlgoLined
\caption{Training one-vs-all SVMs}
\SetKwInOut{Input}{Input}\SetKwInOut{Output}{Output}
\Input{$(\xsi, \ysi) \in (X, Y)$}
\Output{$svm$, a set of 26 SVMs}

\For{$y \leftarrow 0, ..., 26$}{
    $X_{pos} \leftarrow \xsi, \forall \ysi = y$ \\
    $X_{neg} \leftarrow \xsi, \forall \ysi \ne y$ \\
    $svm_y \leftarrow TrainSVM(X_{pos}, X_{neg})$
}
yield $svm$
\end{algorithm}

At prediction time, produce a score from each SVM, select the one that generates the most positive score as the prediction.

\begin{algorithm}[H]
\SetAlgoLined
\caption{Predicting with one-vs-all SVMs}
\SetKwInOut{Input}{Input}\SetKwInOut{Output}{Output}
\Input{$svm$, a set of 26 SVMs; $\hat{x}$, an unseen sample}
\Output{$\hat{y}$, a predicted label}

yield $\arg \max_i SVMPredict(svm_i, \hat{x})$
\end{algorithm}

\subsection{Convolutional Neural Network (CNN)}

The format of our interpolated, re-sampled data consists of three independent components, yaw, pitch, and roll, where each has identical shape of $(NumFeatures,)$. We may use a CNN\cite{lenet} to learn from this data by considering each component to be in its individual channel, and construct layers of 1-D convolution, with activation and pooling to construct a network. After experimentation, our best performing network has the following structure:

\begin{enumerate}[label=-]
    \item Input: 3 channels of $100 \times 1$ vectors
    \item 3 Convolution $\to$ max-pool $\to$ ReLU: number of channel increases $3 \to 32\to 64$, kernel size 3 stride 1 with padding, size $100 \times 1$ unchanged 
    \item 3 Fully connected layers of width $6400 \to 3200 \to 1600 \to 500$ with ReLU activation
    \item Output: vector of $26\times 1 $ of one-hot encoding for each letter.
\end{enumerate}

The training label (and subsequently prediction output) is a one-hot encoding of each of the 26 letters. We train the CNN by gradient descent, minimizing cross entropy:
\[
J(x, y) = - \sum_{i=1}^n \sum_{c=1}^C y_c^{(i)} \log \hat{y}_c^{(i)}
\]

From observing the training result of the two baseline models (KNN and SVM), we realize that they result in very low test accuracy using subject split, meaning that with a more powerful model, we need to be careful to not overfit to the training set. Therefore, in training the CNN, we use aggressive L2 regularization with $\lambda = 0.1$ when calculating gloss. 

\subsection{Recurrent Neural Network (RNN)}

One of the state-of-the-art models for representing sequential data is Recurrent Neural Network (RNN). Unlike feedforward networks such as CNNs, RNNs consider inputs in a sequential data at each time step as an individual node and build connections among the nodes to form a directed graph in the temporal manner. Such structure has the inherent benefits for computing sequential data as it is able to utilize its intermediate hidden states as "memories" to store information about inputs in the previous time stamps.

Long Short-Term Memory (LSTM)\cite{lstm} is a specific type of RNN consisting of three components: input gates, output gates, forget gates. Input gates handle the new data fed into the network, output gates determine the values stores in the cells to be used to calculate the output activation, and forget gates control the values to be kept in the cells.

In this project, we designed a 5-layer LSTM cell as our RNN model to tackle this problem. For each node in the input sequence, each layer computes the following:
\[
\begin{array}{ll} 
    i_t = \sigma(W_{ii} x_t + b_{ii} + W_{hi} h_{(t-1)} + b_{hi}) \\
    f_t = \sigma(W_{if} x_t + b_{if} + W_{hf} h_{(t-1)} + b_{hf}) \\
    g_t = \tanh(W_{ig} x_t + b_{ig} + W_{hg} h_{(t-1)} + b_{hg}) \\
    o_t = \sigma(W_{io} x_t + b_{io} + W_{ho} h_{(t-1)} + b_{ho}) \\
    c_t = f_t * c_{(t-1)} + i_t * g_t \\
    h_t = o_t * \tanh(c_t) \\
\end{array}
\]

where $i_t$ is the input gate at time $t$, $f_t$ is the forget gate at time $t$ and $o_t$ is the output gate at time $t$. $g_t$ is an intermediate state that filters and activates the long term memory. $h_t$ is the hidden state, or short term memory, at time $t$, $c_t$ is the cell state at time $t$ and $x_t$ is the input at time $t$. $\sigma$ is the sigmoid function, and $*$ is the Hadamard product. $W_{jk}$ and $b_{jk}$ are weights and bias for the filter connecting gate $j$ and gate $k$.

For our 5-layer LSTM, the input for the $i$th layer is the product hidden states of the $i-1$th layer for all layers except the first layer. We randomly initialize the hidden states and the cell state with a standard normal distribution. Cross-entropy loss is also used to train the RNN model.

\section{Experiments}

\subsection{Splitting Data for Training and Testing}

We propose two ways to split the data set for training.

\textbf{Classic random split}: randomly split the data set with proportion to 80:10:10 for train set, dev set, and test set respectively. We expect this would train the model to learn writing pattern from each subject then make a prediction on the seen subjects.

\textbf{Individual subject split}: randomly choose two subjects for dev, two subjects for test, and use all other subjects' data for training. We expect this would lead to lower performance, but indicate how well each model generalizes to the overall population.

Because we have a small amount of data, ideally it is more appropriate to use cross-validation as the method to assess the generality and accuracy of the models. However, due to the time consuming nature of RNN training, we were not able to complete cross-validation for a large enough number of folds. As a result, the accuracy reported below were calculated using the splitting strategies described above, and we plan on experiment with cross-validation in the future.

\subsection{Data Preprocessing}
\subsubsection{Interpolation and Re-sampling}
\begin{figure}[H]
    \centering
    \includegraphics[scale = 0.21]{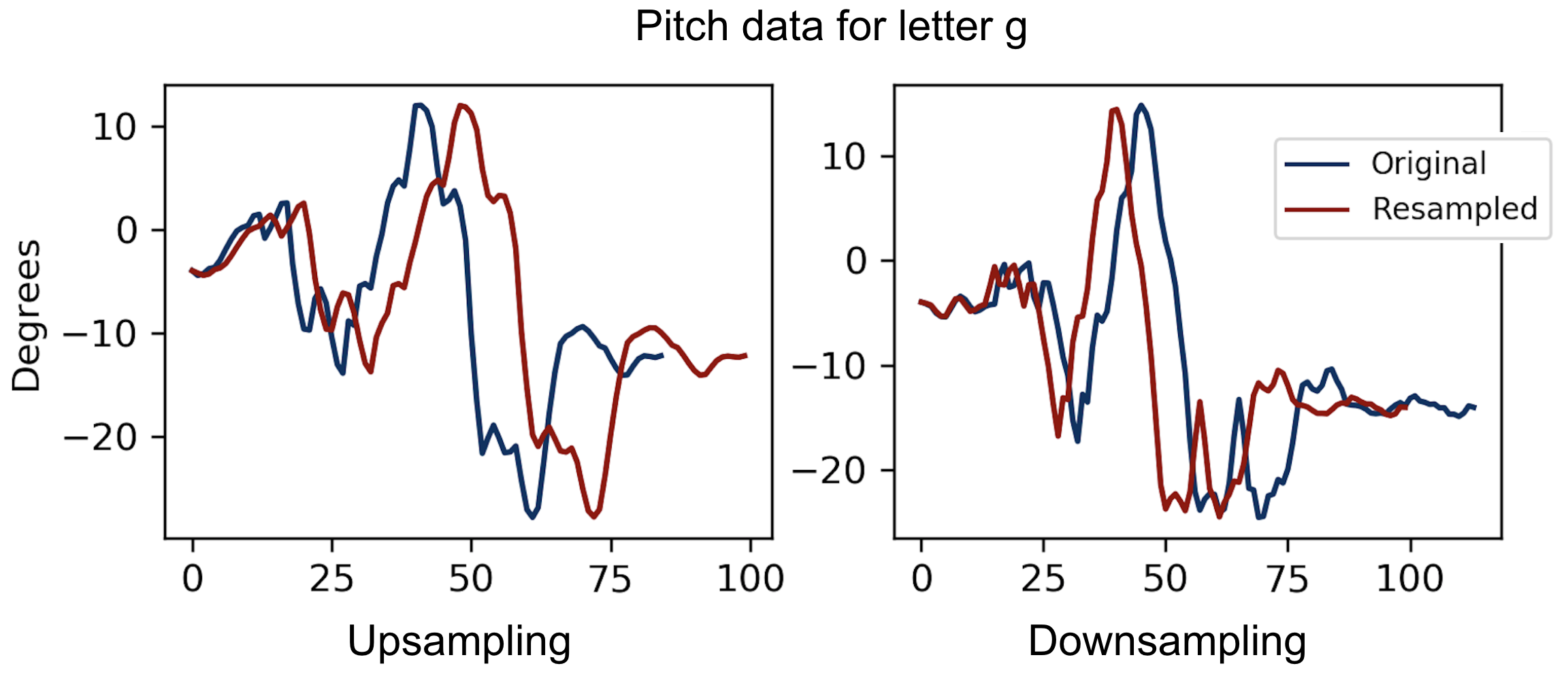}
    \caption{Compare before and after re-sampling}
\end{figure}
Due to the different writing habits among subjects, the data sequence we collected do not have a consistent length by nature. By setting our fixed feature number to be 100, we downsample data from those who write at a slower pace, and upsample data from people who write faster, using the interpolation technique introduced in section 3.4, and an example visualization of the result is shown as above.

\subsubsection{Data Augmentation}
\begin{figure}[H]
    \raggedleft
    \begin{subfigure}[]{0.25\textwidth}
        \raggedleft
        \includegraphics[scale = 0.35]{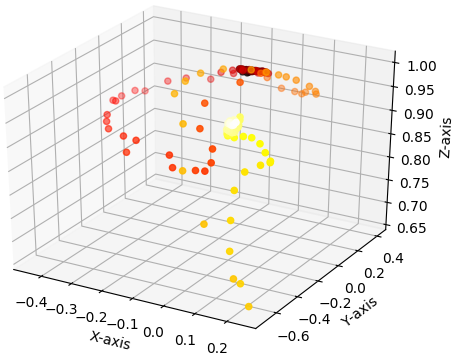}
        \caption{A data sequence for letter "a" without augmentation.}
    \end{subfigure}%
    ~ 
    \begin{subfigure}[]{0.25\textwidth}
        \centering
        \includegraphics[scale = 0.35]{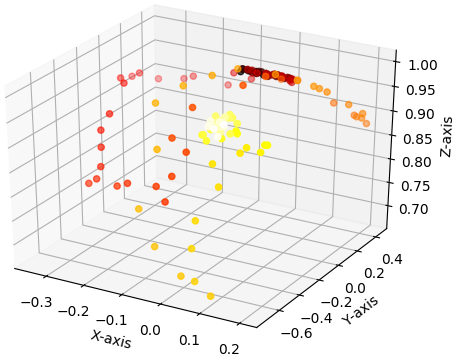}
        \caption{A data sequence for letter "a" with augmentation.}
    \end{subfigure}
    \caption{Data Augmentation Demonstration}
\end{figure}

Observe that after augmenting the sequence, the pattern is similar to the original and distinct from sequences of other labels.

\subsubsection{Denoising Autoencoder}
\begin{figure}[H]
    \centering
    \includegraphics[scale = 0.21]{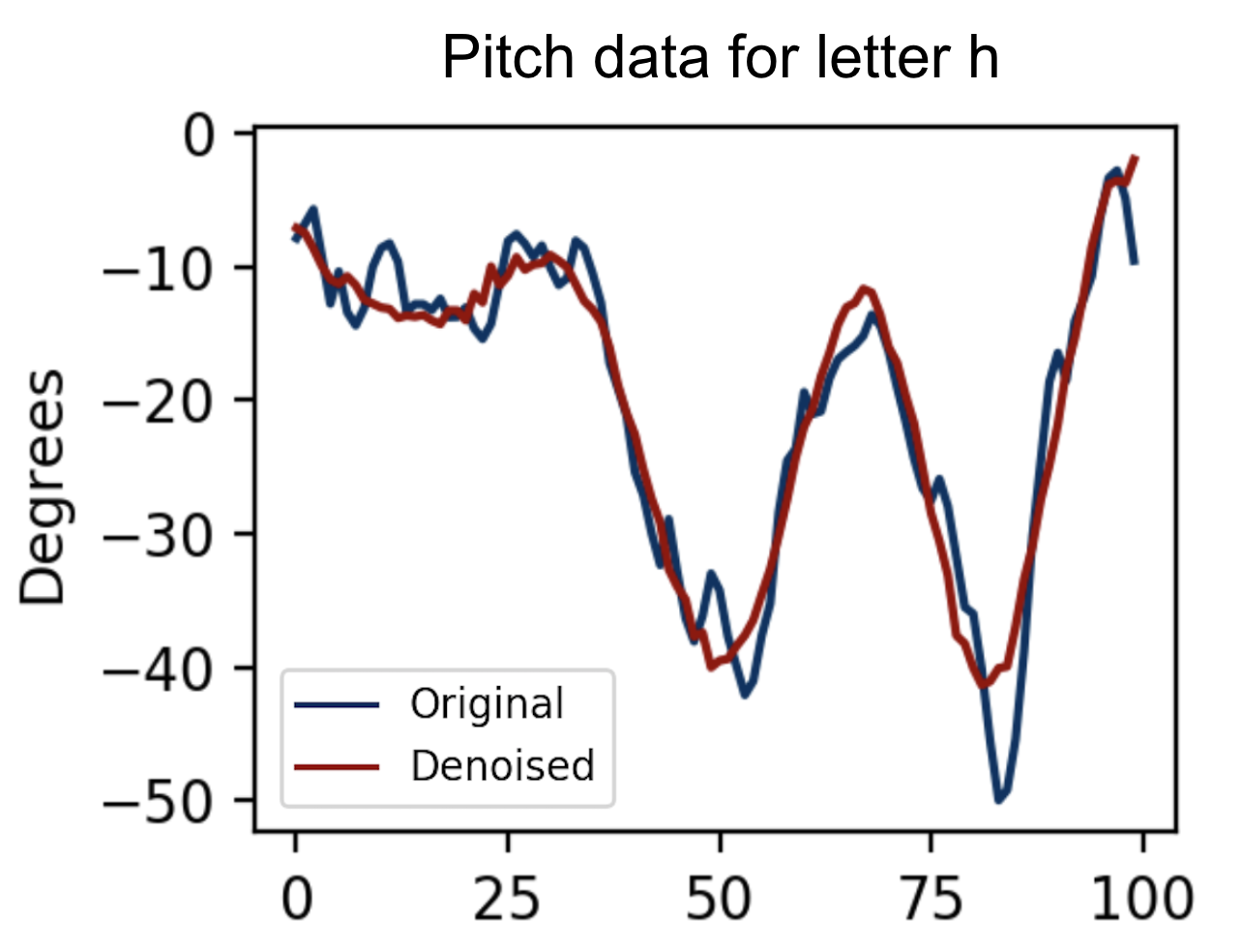}
    \caption{A data sequence for letter "a" without augmentation.}
\end{figure}

\vspace{-15px}

The autoencoder network appears to learn to encode the descriptive features of the raw input, while discarding noise from the raw sequence, to produce a smooth curve that still very much resembles the original.

\subsection{Baseline Models}

\begin{table}[H]
\centering
\begin{tabular}{lllllll}
\hline
K           & 2    & 3    & 4    & 5    & 6    & 7    \\ \hline
Rand. split & .721 & .718 & .718 & .707 & .697 & .693 \\
Subj. split & .126 & .127 & .128 & .126 & .129 & .131 \\ \hline
\end{tabular}
\caption{KNN Test Accuracy}
\label{tab:knn-acc}
\end{table}

\vspace{-14px}

\begin{table}[H]
\centering
\begin{tabular}{lll}
\hline
      & Random split & Subject Split \\ \hline
Train & 0.753125     & 0.9998        \\
Test  & 0.590778     & 0.15562       \\ \hline
\end{tabular}
\vspace{3pt}
\caption{SVM Accuracy}
\label{tab:svm-acc}
\end{table}

\vspace{-11px}

We observe that neither KNN nor SVM achieve decent accuracy with subject split, but is able to generalize relatively well with classic random split, with KNN achieving 0.721 test accuracy at $K=2$.

\subsection{CNN}

The model is given a computation budget of 20 epochs, trained with a mini-batch gradient descent with batch size 500. We use Adam optimization\cite{adam} to train the model. The model shows the following convergence behavior.

\vspace{-10px}
\begin{figure}[H]
    \centering
    \includegraphics[scale=0.35]{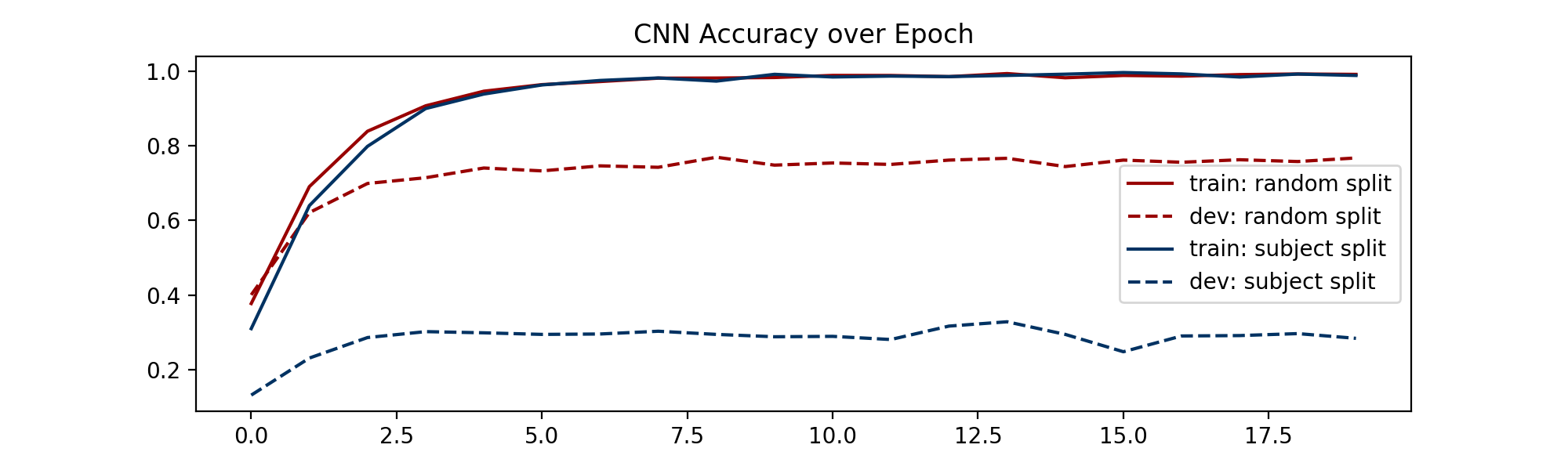}
    \caption{CNN Training and Validation Accuracy}
    \label{fig:cnn_train_acc_plot}
\end{figure}

\vspace{-12px}

We observe that for both splits, the model is able to fit well to the training set, but only generalizes well to the validation set with random split, even with aggressive L2 regularization. 

\subsection{RNN}

\begin{figure}[H]
    \centering
    \includegraphics[scale=0.35]{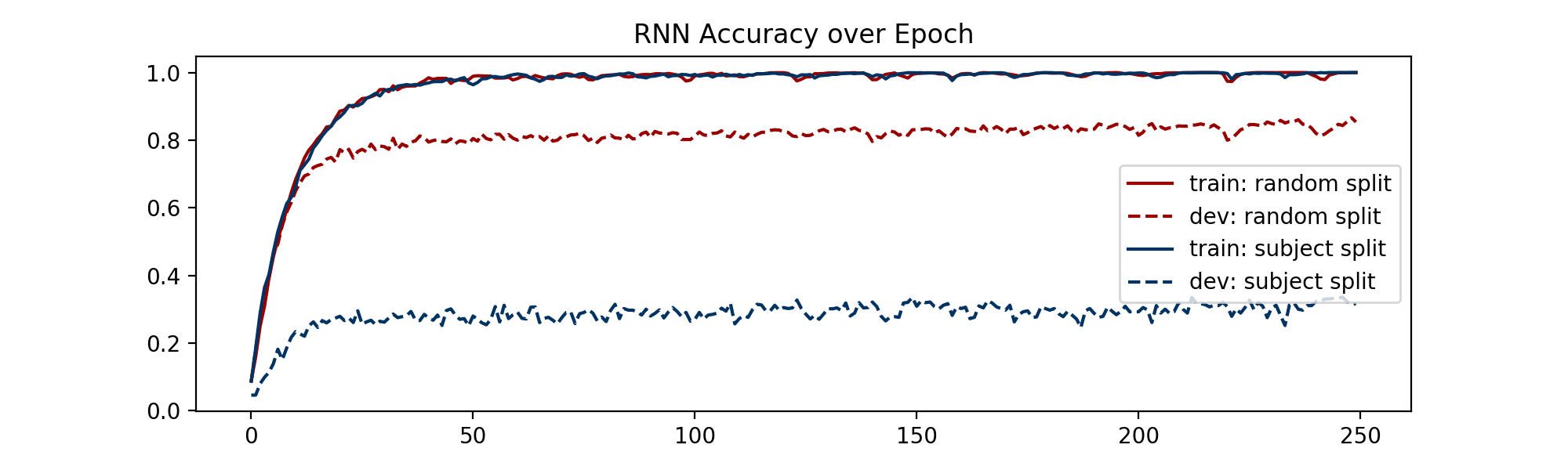}
    \caption{RNN Training and Validation Accuracy}
    \label{fig:rnn_train_acc_plot}
\end{figure}

We use the same mini-batch gradient descent strategy with feature interpolation and without data flatten to train the RNN model. In comparison to CNN, RNN takes many more epochs before it converges, and therefore we give it a computation budget of 250. We observe that validation accuracy shows improvement over its CNN counterpart in both random split and subject split. However, it still suffers from severe overfitting problem with subject pslit experiment given the large gap between the training and validation curve.

\subsection{Ablation Study}
\subsubsection{Data Augmentation}



\begin{table}[H]
\centering
\begin{tabular}{llllll}
                               &                  & \multicolumn{2}{l}{CNN} & \multicolumn{2}{l}{RNN} \\ \hline
                               &                  & Train      & Test       & Train      & Test       \\ \hline
\multirow{2}{*}{Random Split}  & w/o aug & 0.985      & 0.710      & 1.000      & 0.766      \\ 
                               & w/ aug  & 0.919      & \textbf{0.774}      & 0.999      & \textbf{0.843}      \\ \hline
\multirow{2}{*}{Subject Split} & w/o aug & 0.977      & 0.358      & 1.000      & 0.406      \\  
                               & w/ aug  & 0.988      & \textbf{0.396}      & 1.000      & \textbf{0.510}      \\ \hline
\end{tabular}
\vspace{3pt}
\caption{Accuracy with and without Data Augmentation}
\label{tab:data-aug-acc}
\end{table}

\vspace{-15px}
Table \ref{tab:data-aug-acc} highlights the difference between deep models trained with and without data augmentation. As we can observe, the data augmentation method is very effective to improve the test accuracy while all models converge at high training accuracy. Such observation proves our statement about model overfitting problem that we concluded in the previous sections, since data augmentation increases the diversity of the training dataset to help generalization.


\subsubsection{Data Denoising}
\begin{table}[H]
\centering
\begin{tabular}{llllll}
                               &        & \multicolumn{2}{l}{CNN} & \multicolumn{2}{l}{RNN} \\ \hline
                               &        & Train      & Test       & Train      & Test       \\ \hline
\multirow{2}{*}{Random Split}  & w/o AE & 0.919      & 0.774      & 0.999      & 0.843      \\  
                               & w/ AE  & 0.996      & \textbf{0.784}      & 0.999      & \textbf{0.866}      \\ \hline
\multirow{2}{*}{Subject Split} & w/o AE & 0.988      & 0.396      & 1.000      & 0.510      \\  
                               & w/ AE  & 0.996      & \textbf{0.402}      & 0.999      & \textbf{0.536}      \\ \hline
\end{tabular}
\vspace{3pt}
\caption{Accuracy with and without Denoising Autoencoder}
\label{tab:denoise-ae-acc}
\end{table}

\vspace{-15px}

The above result shows that denoising the data with a trained autoencoder results in higher accuracy with both network across both splits of the dataset. This affirms our hypothesis that small fluctuations in raw sensor data are indeed noise that are independent from the writing pattern, and that removing them allows the classifier to better learn the descriptive features of each class against others.

\subsection{Overall Performance and Discussion}

\begin{table}[H]
\centering
\begin{tabular}{lllll}
\hline
            & \multicolumn{2}{l}{Random split} & \multicolumn{2}{l}{Subject split} \\ 
            & Train           & Test           & Train           & Test            \\ \hline
KNN(K=4)    & -               & 0.718          & -               & 0.128           \\
SVM         & 0.753           & 0.591          & 0.999           & 0.156           \\
CNN         & 0.985           & 0.710          & 0.977           & 0.358           \\
CNN(aug)    & 0.919           & 0.774          & 0.988           & 0.396           \\
CNN(aug+AE) & 0.996           & 0.784          & 0.996           & 0.402           \\
RNN         & \textbf{1.000}           & 0.776          & \textbf{1.000}           & 0.406           \\
RNN(aug)    & 0.999           & 0.843          & \textbf{1.000}           & 0.510           \\
RNN(aug+AE) & 0.999           & \textbf{0.866}          & 0.999           & \textbf{0.536}           \\ \hline
\end{tabular}
\vspace{3pt}
\caption{Overall Accuracy}
\label{tab:overall-acc}
\end{table}

\vspace{-15pt}

With many samples and complex patterns, our baseline model, SVM, failed to converge within the computation budget of 40000 iterations. This is evidence that an SVM with a polynomial kernel is not descriptive enough to learn the complex pattern. KNN, on the other hand, performs well with a classic random split, meaning that it is able to find a number of neighbors that are similar, but fail to generalize to the population, leading to low accuracy with subject split. 

Our RNN with data augmentation and autoencoder denoising perform the best, both because having more, less noisy data samples result in better learning of features, but also because RNN is able to learn features from a time series, resulting in an advantage over CNNs.

We observe that all models suffer from overfitting to the training data. We hypothesize that this is not over-parameterization, as the networks are simple with few parameters in comparison to input features. By observing accuracy in subject split, we notice test accuracy fluctuates substantially as different runs select different test subjects, and accuracy decreases as the writing habits of test subjects deviate from ones of train subjects. From this we interject that there are a few reasons for the low accuracy:

\begin{enumerate}[label=-]
    \item Writing habit differs greatly across the population
    \item Rotation sensor data alone may not be distinctive enough
    \item The lowercase English alphabet has many similar letters that obfuscate the classifier
\end{enumerate}

\section{Conclusion, Future Work}

We reach the conclusion based on our technique and experiments that, albeit having a noisy small dataset, it is possible to achieve high accuracy in handwriting recognition based on rotation sensor data, given the user calibrates the model with its handwriting habits before it makes predictions. 

For future work, we plan to explore additional way to approach motion-based recognition that are either more plausible to achieve high accuracy, or better generalize to the population with limited amount of data:

\begin{enumerate}[label=-]
    \item Categorize on uppercase English alphabet, as they are far more distinct than lowercase letters;
    \item Gesture recognition, as users are free to define their own gestures, and that they are by definition more distinct among each other;
    \item Remove the pen limitation and use other more intuitive / more accurate sensors, such as a ring, or a VR controller;
    \item Use cross-validation to obtain more accurate representation of the model performance;
    \item Report confusion matrix for the entire alphabet.
\end{enumerate}





\bibliographystyle{unsrt}  
\bibliography{reference}

\end{multicols*}
\end{document}